\documentclass{article}

\usepackage{arxiv}

\usepackage[utf8]{inputenc} 
\usepackage[T1]{fontenc}    
\usepackage{hyperref}       
\usepackage{url}            
\usepackage{booktabs}       
\usepackage{amsfonts}       
\usepackage{nicefrac}       
\usepackage{microtype}      
\usepackage{lipsum}		
\usepackage{graphicx}
\usepackage{natbib}
\usepackage{doi}
\usepackage{subcaption}
\usepackage{amsmath}

\title{Modeling chaotic Lorenz ODE System using Scientific Machine Learning}

\author{ \href{https://orcid.org/0009-0007-3451-9352}{\includegraphics[scale=0.06]{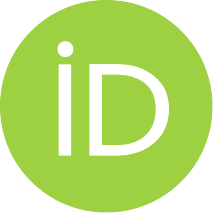}\hspace{1mm}Sameera S Kashyap} \\
	Acko\\
	\texttt{sameera.s.kashyap@gmail.com} \\
	\And
	\href{https://orcid.org/0000-0003-1946-0667}{\includegraphics[scale=0.06]{orcid.pdf}\hspace{1mm}Raj Abhijit Dandekar} \\
	Vizuara AI Labs\\
        Massachusetts Institute of Technology (prior)\\
	\texttt{raj@vizuara.com} \\
 \And
	\hspace{1mm}{Rajat Dandekar} \\
	Vizuara AI Labs\\
       Purdue University (prior)\\
	\texttt{rajatdandekar@vizuara.com} \\
 \And
	\hspace{1mm}{Sreedath Panat} \\
	Vizuara AI Labs\\
         Massachusetts Institute of Technology (prior)\\
	\texttt{sreedath@vizuara.com} \\
}

\hypersetup{
pdftitle={A template for the arxiv style},
pdfsubject={q-bio.NC, q-bio.QM},
pdfauthor={David S.~Hippocampus, Elias D.~Striatum},
pdfkeywords={First keyword, Second keyword, More},
}

\begin{document}
\maketitle

\begin{abstract}
	In climate science, models for global warming and weather prediction face significant challenges due to the limited availability of high-quality data and the difficulty in obtaining it, making data efficiency crucial. In the past few years, Scientific Machine Learning (SciML) models have gained tremendous traction as they can be trained in a data-efficient manner, making them highly suitable for real-world climate applications. Despite this, very little attention has been paid to chaotic climate system modeling utilizing SciML methods. In this paper, we have integrated SciML methods into foundational weather models, where we have enhanced large-scale climate predictions with a physics-informed approach that achieves high accuracy with reduced data. We successfully demonstrate that by combining the interpretability of physical climate models with the computational power of neural networks, SciML models can prove to be a reliable tool for modeling climate. This indicates a shift from the traditional black box-based machine learning modeling of climate systems to physics-informed decision-making, leading to effective climate policy implementation.


\end{abstract}


\section{Introduction}

The Lorenz system of equations is a set of ordinary differential equations to represent a simplified model of atmospheric convection \cite{sparrow1982lorenz}. These set of equations have a wide range of applications in fields ranging from fluid mechanics to laser physics to weather prediction. 

One of the most interesting properties of the Lorenz ODE System is that it is chaotic in nature \cite{fowler1982complex}. Small changes in the initial conditions can lead to vastly different outcomes in the end result \cite{lorenzMathematics}. When simulated over a given period, the Lorenz ODEs show oscillations in time. Usually, numerical methods implemented in computational software modeling tools like Python, Julia, or Matlab are used to simulate the Lorenz System of ODEs. These methods are inefficient as Lorentz equations are sensitive to initial conditions and minute changes to the conditions and tiny rounding errors can lead to the accumulation of numerical errors over time. Very few studies have been aimed at integrating machine learning-aided methods in simulating the chaotic Lorenz system. In this study, we provide a robust investigation of the effect of two physics-aided machine learning models in simulating the Lorenz system of ODEs: Neural Ordinary Differential Equations (Neural ODEs) \cite{node1} and Universal Differential Equations (UDEs)
\cite{ude1}.

\subsection{Impact on climate modeling}

In our study, we demonstrate how these Scientific ML models can be integrated into one of the fundamental models of weather prediction: The Lorenz ODEs. This opens the door to integrating these methods into large-scale weather prediction models, which would traditionally require a large amount of data \cite{weather1, weather2}. One of the key advantages of Scientific ML methods, such as Neural ODEs and UDEs, is their ability to encode underlying physical laws into the model. This physics-informed approach allows these models to achieve high accuracy with less data compared to traditional machine-learning models. In the context of climate science models such as global warming or weather prediction models, where high-quality data can be sparse or difficult to obtain, this data efficiency is crucial. By requiring less data, SciML models can be trained more quickly and with fewer resources, making them highly suitable for real-world climate applications where rapid and accurate predictions are essential \cite{weather3}.

As governments and organizations seek to implement effective climate strategies, we need to look at integrating the interpretability of physical climate models with the power of neural networks. As we demonstrate in our study, the ability of Neural ODEs and UDEs to model and forecast chaotic systems with high accuracy ensures that policymakers have access to reliable information, empowering them to make informed decisions that address the climate crisis.

\subsection{Neural ordinary differential equations and universal differential equations}

Neural Ordinary Differential Equations and Universal Differential Equations come under the pillar of the rapidly growing field of Scientific Machine Learning \cite{node1, ude1}. Scientific Machine Learning (SciML) merges physical models with neural networks. In doing so, this field of SciML leverages the interpretability of scientific structures with the expressivity of neural networks. This combination makes the SciML approach very powerful in discovering previously unknown physics and solving complex differential equations with even less data. As of this writing, there have been multiple studies investigating the applications of Scientific Machine Learning in the fields of epidemiology, optics, gene modeling, civil engineering, fluid mechanics, battery modeling, material discovery, drug discovery, quantum circuits, astronomy and a huge range of related fields \cite{scimlappl1, scimlappl2, scimlappl3, scimlappl4, scimlappl5, scimlappl6, scimlappl7, scimlappl8, scimlappl9, scimlappl10, scimlappl11, scimlappl12, scimlappl13, scimlappl14}. SciML uses methodologies that combine traditional ODE and PDE solvers with neural networks, where the network partially or entirely replaces the equation, the neural network is then trained using gradient descent to obtain a model that captures the underlying physics. Broadly, the rise of Scientific Machine Learning can be attributed to three popular methodologies:

\begin{itemize}
    \item Neural Ordinary Differential Equations (Neural ODEs): The entire forward pass of an ODE/PDE is replaced with neural networks. We perform backpropagation through the neural network augmented ODE/PDE. In doing so, we find the optimal values of the neural network parameters. \cite{node1, node2, node3, node4}
    \item Universal Differential Equations (UDEs): In contrast to Neural ODEs, only certain terms of the ODE/PDEs are replaced with neural networks. We then discover these terms by optimizing the neural network parameters. Universal Differential Equations can be used to correct existing underlying ODEs/PDEs as well as to discover new, missing physics. \cite{ude1, ude2, ude3, ude4}
    \item Physics Informed Neural Networks (PINNs): PINNs are predominantly used as an alternative to traditional ODE/PDE solvers to solve an entire ODE/PDE. We replace the output variable with a neural network and the loss function is determined by the ODE/PDE solution and the boundary conditions. When we minimize the loss function, we automatically find the optimal solution to the ODE/PDE. \cite{pinn1, pinn2, pinn3, pinn4}
\end{itemize}

Despite the rise in Scientific Machine Learning frameworks, very little attention has been paid to the systematic applications of the above SciML pillars on chaotic ODE system modeling, like the Lorenz system. Although some studies have investigated Neural ODEs applications on the Lorenz system \cite{climatesciml1, climatesciml2, climatesciml3, climatesciml4}, there is no study that has looked at integrating Universal Differential Equations (UDEs) with the Lorenz system. In particular, the following questions are still unanswered:

\begin{itemize}
    \item Can Neural ODEs and UDEs be used as alternatives for traditional system modeling tasks involving numerical differentiation?
    \item In the spirit of UDEs, can we replace certain terms of the Lorenz system with neural networks and recover them?
    \item How does the Neural ODE prediction compare with the UDE prediction?
    \item Can we do forecasting on the Lorenz system ODEs with Neural ODEs and UDEs?
    \item Are UDEs better at forecasting than Neural ODEs?
\end{itemize}

The main goal of our study is to answer the above 4 questions. In our study, we construct a detailed methodology for implementing the Neural ODE and UDEs for the Lorenz System. We use the advanced Scientific Machine Learning libraries provided by the Julia Programming Language \cite{Julia1, Julia2, Julia3, Julia4}. Through robust hyperparameter optimization testing, we provide insights into the neural network architecture, activation functions, and optimizers that provide the best results. We show that both Neural ODEs and UDEs can be used effectively for both prediction and forecasting of the Lorenz ODE system. More importantly, we introduce the "forecasting breakdown point" - the time at which forecasting fails for both Neural ODEs and UDEs. This provides an insight into the applicability of Scientific Machine Learning frameworks in forecasting tasks. 

The paper is structured as follows. We start by presenting the methodology and detailed description for Neural ODEs and UDEs and we adapt these methodologies for the Lorenz system. Subsequently, we present the prediction and forecasting results for the Neural ODEs and UDEs. We also provide hyperparameter optimization plots, concluding with limitations describing the forecasting breakpoint and the effect of noise on training and forecasting on the model. 

\section{Methodology}

This section describes the methodological approach to exploring chaotic dynamics in climate modeling through Neural ODEs and UDEs on the Lorenz system.

The equations are defined as:
\begin{equation}\frac{dx}{dt} = \sigma (y - x)\end{equation}
\begin{equation}\frac{dy}{dt} = x (\rho - z) - y\end{equation}
\begin{equation}\frac{dz}{dt} = xy - \beta z\end{equation}

The constants \textbf{$\sigma$}, \textbf{$\rho$}, and \textbf{$\beta$} are specifically used in the Lorenz system because they are derived from the physical parameters of atmospheric convection and help to capture the essence of the convective process in a simplified mathematical model.

\textbf{$\sigma$ - The Prandtl Number}: \cite{Prandtl} represents the ratio of the rate of momentum diffusion to thermal diffusion. In the context of the Lorenz system, 
$\sigma$ = 10 is a common value chosen to reflect the properties of air, which has a Prandtl number around this value. This value helps to ensure that the model accurately captures the relative effects of viscosity and thermal diffusion in atmospheric convection

\textbf{$\rho$ - The Rayleigh Number}: \cite{Rayleigh} It quantifies the driving force for convection due to the temperature difference across the fluid layer. The value 
$\rho$ = 28 is chosen to place the system in a chaotic regime. In the Lorenz system, this particular value is known to lead to chaotic behavior, which is of interest for studying complex, unpredictable dynamics similar to those observed in the atmosphere.

\textbf{$\beta$ - The Geometric Factor}: It accounts for the aspect ratio of the convective cells. The specific value $\beta$ = 8/3 is chosen based on empirical and theoretical studies of convective systems. This value ensures that the simplified model retains key characteristics of the full convection equations, allowing it to accurately represent the interplay between the fluid's horizontal and vertical motions.

\begin{figure}[h]
    \centering
    \begin{subfigure}[b]{0.45\textwidth}
        \centering
        \includegraphics[width=\textwidth]{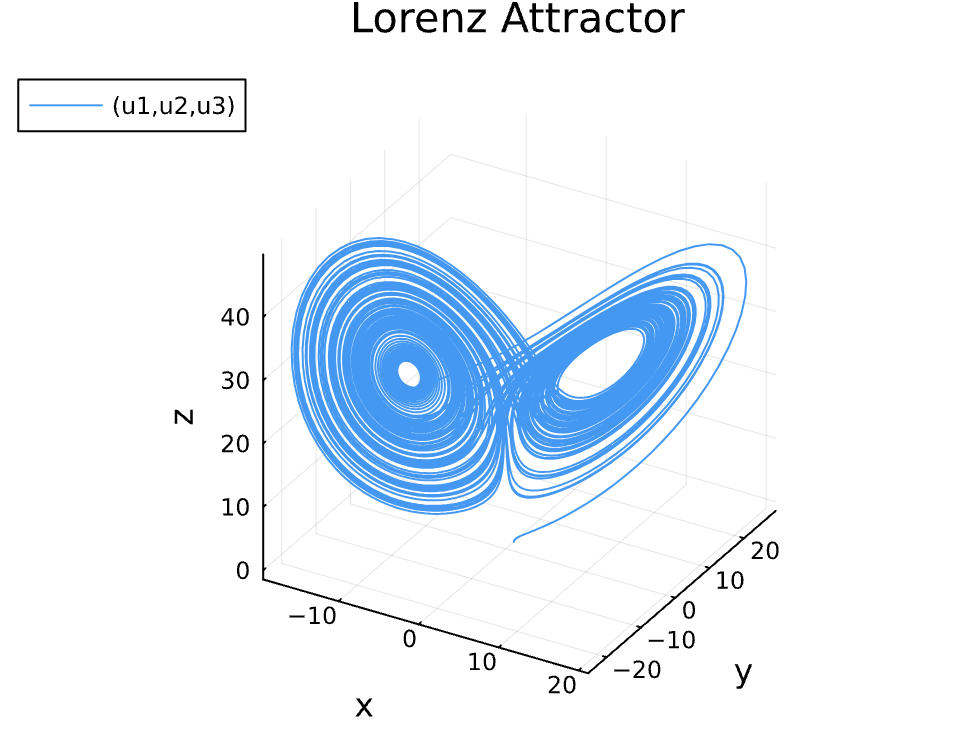}
        \caption{The Butterfly Effect.}
        \label{fig:fig1}
    \end{subfigure}
    \hfill
    \begin{subfigure}[b]{0.45\textwidth}
        \centering
        \includegraphics[width=\textwidth]{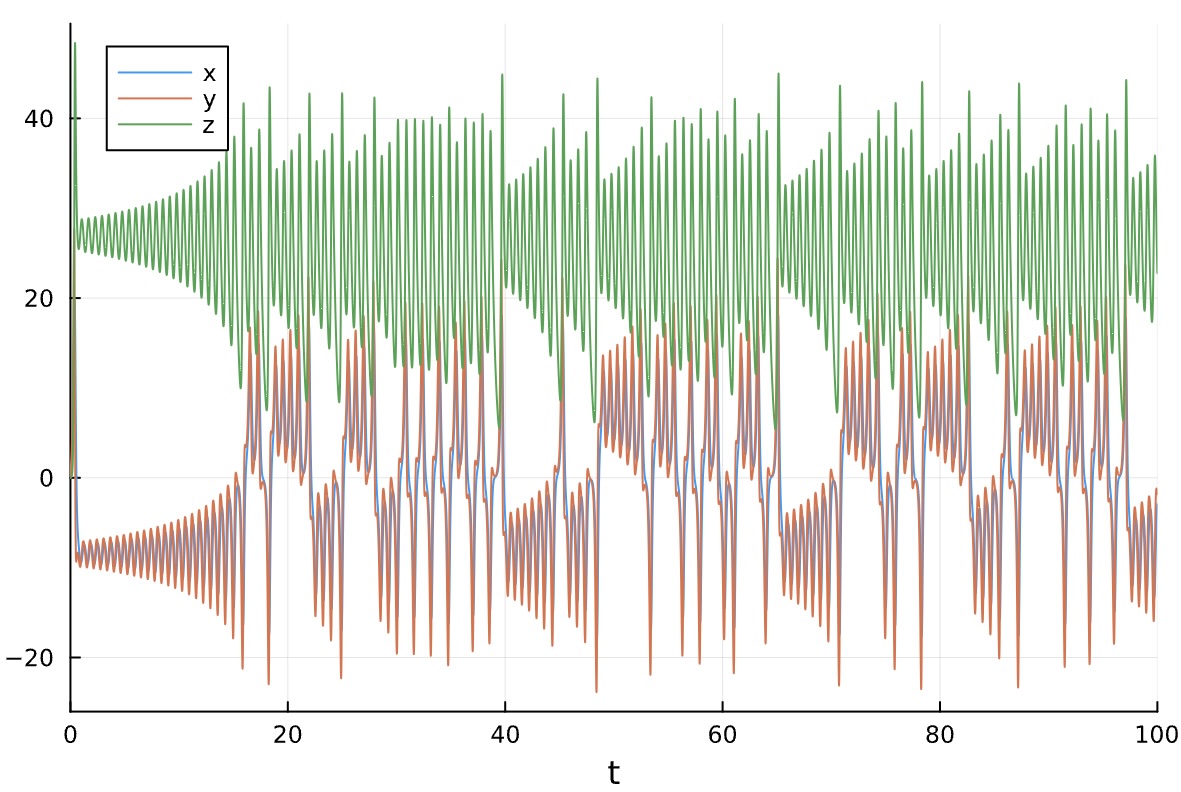}
        \caption{The Lorenz attractor}
        \label{fig:fig2}
    \end{subfigure}
    \caption{Plots showcasing the solution to the chaotic Lorenz ODE system of equations which generate a pair of spiral swirls when projected in three dimensions.}
    \label{fig:lorenz}
\end{figure}

As shown in Figure \ref{fig:lorenz}, the chaotic solutions of the Lorenz system often settle into a pattern known as the Lorenz attractor. The Lorenz attractor is a fractal structure in phase space that illustrates the complex, deterministic yet unpredictable behavior of the system. It consists of two lobes, where the system's trajectory oscillates back and forth in an aperiodic manner, never repeating exactly but confined to a bounded region. This is exactly the outcome that is produced on solving the Lorenz system of equations using ordinary differential equations, it is especially useful for the inherent unpredictability in weather systems due to sensitivity to initial conditions, a concept often summarized by the "butterfly effect."

For all experiments conducted forward, the constants \textbf{$\sigma$}, \textbf{$\rho$,} and \textbf{$\beta$}, are set to \textbf{\textit{10.0, 28.0 and 8/3}} respectively. These constants control different aspects of the convective motion in the Lorenz system, which was originally developed to model atmospheric convection. these values specify a particular regime of chaotic behavior in the Lorenz attractor, which is useful for studying chaotic dynamics in climate modeling. Now, we will look at different methods for solving the equations using neural networks.

\subsection{Neural ODE}

Neural Ordinary Differential Equations (Neural ODEs) are a class of models that represent continuous-depth neural networks. Introduced by \cite{node1}, Neural ODEs have opened up new possibilities in modeling continuous processes by using differential equations to define the evolution of hidden states in neural networks \cite{nodeapp1, nodeapp2, nodeapp3, nodeapp4, nodeapp5, nodeapp6}. Neural ODEs represent a novel approach in machine learning and computational modeling that combines neural networks with ordinary differential equations (ODEs). Neural ODEs are a subset of the broader spectrum of scientific machine learning and physics-informed machine learning. The key idea behind Neural ODEs is to use a neural network to approximate the solution of an ODE, thereby allowing for flexible modeling of continuous-time dynamics.

In a traditional neural network, hidden states are updated using discrete layers. In contrast, Neural ODEs use a continuous transformation defined by an ordinary differential equation:

\begin{equation}\frac{dh}{dt} = \textit{f} (h(t), t , \theta)\end{equation}

where,
\begin{itemize}
    \item \textit{h(t)} is the hidden state at time \textit{t}.
    \item \textit{f} is a neural network parameterized by $\theta$.
    \item The hidden state evolves according to the function \textit{f}.
\end{itemize}

Neural ODEs can be implemented by setting up a Neural Network function with a set number of Hidden layers, weights for each layer, and the activation function that would be used to train the network. Once the neural network is set up, training a Neural ODE involves solving the differential equation using numerical ODE solvers (e.g., Runge-Kutta methods). The output at the final time \( t_1 \)  is compared to the target, and gradients are back-propagated through the ODE solver to update the parameters $\theta$. Neural ODEs are more efficient than traditional machine learning as they are most suitable to model complex distributions as they use continuous transformations defined by differential equations \cite{nodeBetter}, hence they retain and integrate physical laws into the learning process by combining data-driven and physics-based models making them more memory efficient to model complex systems or system of equations.

To solve the Lorenz system of equations using Neural ODEs, several crucial steps must be followed. The process begins with setting up an ODE solver capable of integrating the differential equations over the desired time span. This solver must be selected based on its ability to handle stiff equations, ensuring numerical stability and accuracy.

First, we define the true data by solving the Lorenz equations using a conventional ODE solver. This solution provides the ground truth trajectories of the system, which are essential for training the Neural ODE model via backpropagation. The true data will be utilized to compute the loss function by comparing the predicted trajectories from the Neural ODE to these reference trajectories.

Next, we design a neural network that will parameterize the function \textit{f} in the Neural ODE framework. The architecture of this neural network is critical and should include a predefined set of hidden layers. These layers can vary in number and size/The input dimension of the neural network must be specified to match the three variables (x, y, and z) in the Lorenz system. Thus, the input layer should accept a 3-dimensional vector representing the state of the system at a given time.

Several options are available for the activation function, including ReLU, Sigmoid, and Tanh. The choice of activation function significantly impacts the network's ability to model the non-linear dynamics of the Lorenz system. ReLU (Rectified Linear Unit) is often favored for its simplicity and effectiveness in deep networks, helping to mitigate the vanishing gradient problem. Alternatively, the Sigmoid function can be used when a smooth output in the range (0, 1) is desired, though it is prone to saturation issues. Tanh (hyperbolic tangent) is another viable option, providing outputs in the range (-1, 1) and offering non-linearity that can capture the intricate dynamics of the system. A detailed description of tested hyperparameters is shown in Table \ref{tab:hp-ode} and will be explained in Section 3.

\subsection{Universal Differential Equations}

 Universal Differential Equations (UDEs) introduced \cite{ude1}, combine traditional differential equations with machine learning models, such as neural networks, to create a more flexible and powerful tool for modeling complex systems. This approach integrates the robustness of classical differential equations with the adaptability of neural networks, allowing for more accurate and efficient modeling of systems with unknown or partially known dynamics. UDEs offer improved predictive power by combining data-driven approaches with physical laws. This is particularly useful in scenarios where purely data-driven models might overfit or fail to generalize \cite{udeapp1, udeapp2, udeapp3, udeapp4}. The physical laws embedded in UDEs constrain the learning process, ensuring that the model adheres to known scientific principles. Compared to purely data-driven models, UDEs often require fewer data points to achieve high accuracy. The known differential equations provide a strong prior that guides the learning process, reducing the amount of data needed for training. This efficiency makes UDEs suitable for applications with limited data availability.

To implement UDEs in our experiment, we will consider the Lorenz Attractor equations and augment them replacing some terms in the equation with its neural network. Some terms are multiplied by a factor to enhance the training accuracy and improve loss optimization.

We modify equations (1, 2, 3) with, the augmented equations are defined as:
\begin{equation}\frac{dx}{dt} = \sigma (y - \text{$NN_1$(t,x,y,z;$\theta$)})\end{equation}
\begin{equation}\frac{dy}{dt} = - y + 0.1 \text{$NN_2$(t,x,y,z;$\theta$)} \end{equation}
\begin{equation}\frac{dz}{dt} = - \beta + z + 10 \text{$NN_3$(t,x,y,z;$\theta$)}  \end{equation}

To experiment with Universal Differential Equations (UDEs) in a manner similar to Neural ODEs, it is essential to establish an ODE solver that can yield the true solution. This true solution serves as a benchmark for performing backpropagation on the augmented UDE equations. By analyzing the outcomes from training Neural ODEs, we can eliminate the less effective activation functions while maintaining the initial values. This refined approach allows us to implement the neural network.

\section{Results}

\subsection{Neural ODE}

The training process involves iteratively updating the network parameters $\theta$ to minimize the loss function, thus ensuring that the Neural ODE accurately captures the dynamics of the Lorenz system. Table \ref{tab:hp-ode} showcases the different hyperparameters considered for the experiment. In this experiment, we explored various hyperparameters to optimize the network's performance. Our focus was on three key areas: activation function, architecture, and optimizer.

\begin{table}[h]
	\caption{Range of hyper-parameters on training data(ODEs)}
	\centering
	\begin{tabular}{lll}
		\toprule
		\cmidrule(r){1-2}
		Hyperparameter   &Value   &Search Range    \\
		\midrule
          Activation Function    &sigmoid      &ReLU, tanh, sigmoid \\
          Learning Rate    &0.01      &0.1, 0.01, 0.001 \\
          Optimization Solver    &Adam      &Adam, RAdam \\
          Hidden units    &25      &25, 50, 100, 125 \\
          Number of Epochs    &50000      &1000 - 50000 \\
          Loss    &2.15      &2.15, 4.16, 1450, 31075, 90226 \\
		\bottomrule
	\end{tabular}
	\label{tab:hp-ode}
\end{table}

\begin{figure}[h!]
    \centering
    \begin{subfigure}[b]{0.49\textwidth}
        \centering
        \includegraphics[width=\textwidth]{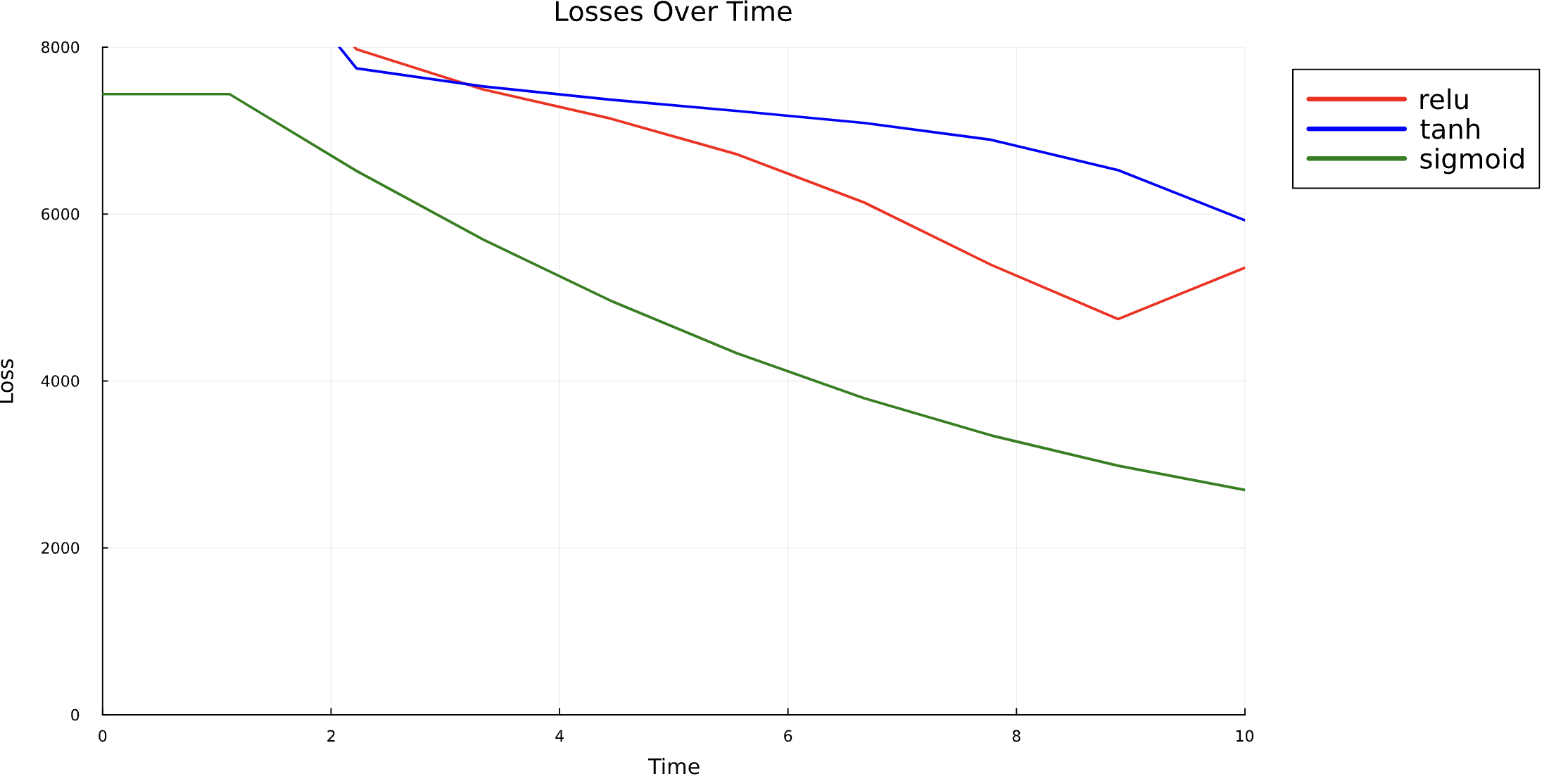}
        \caption{Effect of different activation functions}
        \label{fig:activation_functions}
    \end{subfigure}
    \begin{subfigure}[b]{0.49\textwidth}
        \centering
        \includegraphics[width=\textwidth]{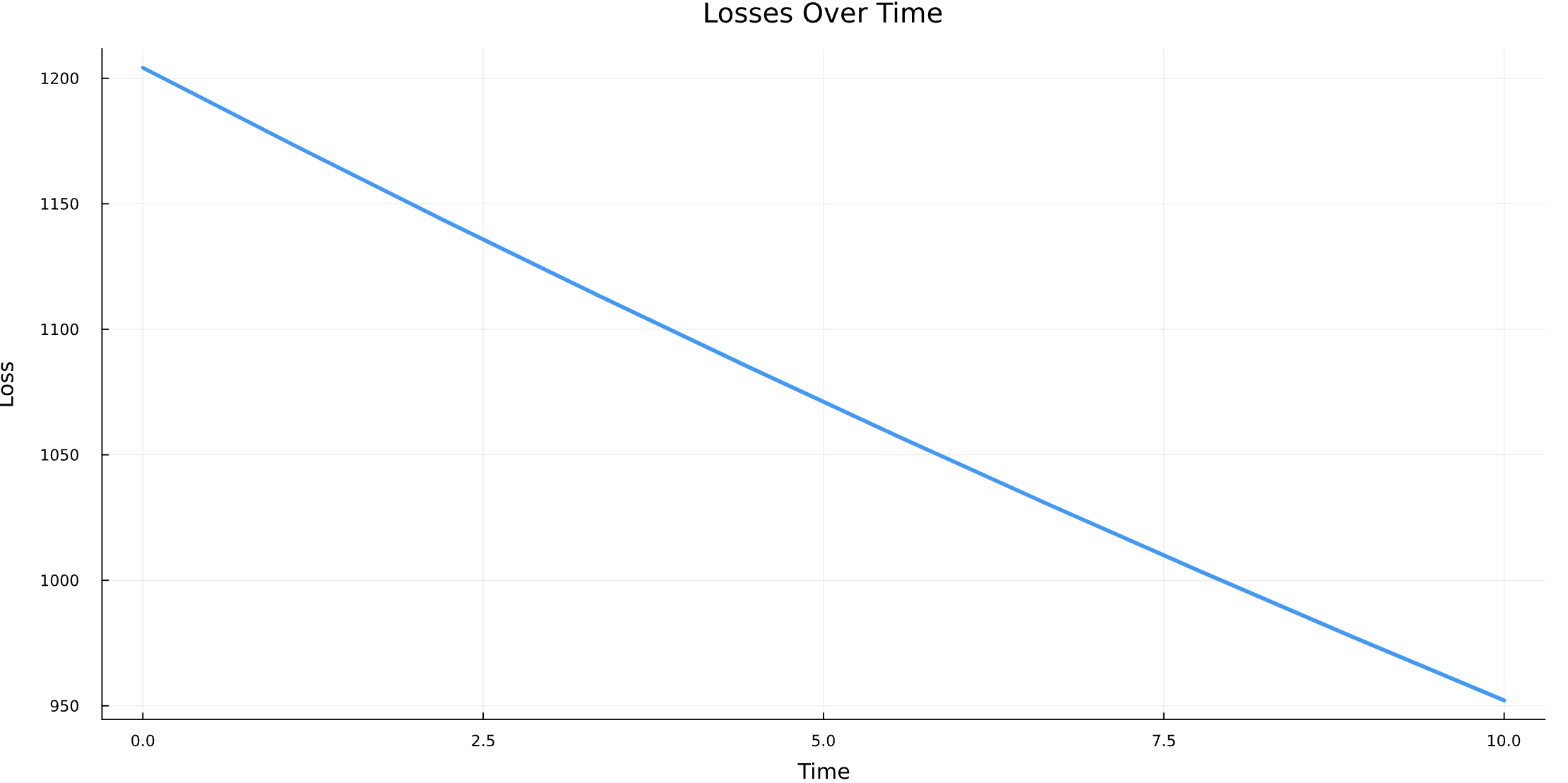}
        \caption{Exponential loss curve of sigmoid}
        \label{fig:ode_losses}
    \end{subfigure}
    \caption{Both \textit{tanh} and \textit{ReLU} struggle to consistently reduce the loss, showing fluctuations but largely stagnating. In contrast, the \textit{sigmoid} activation function resulted in a steadily decreasing, exponential loss curve. The second Figure shows the trend of exponential loss when sigmoid is employed.  }
    \label{fig:both}
\end{figure}

Firstly, we experimented with different activation functions. Among them, the sigmoid function yielded the best results, smoothing nonlinear transformations effectively and enhancing the network's ability to model complex relationships. ReLU, though initially considered, showed stagnant loss values throughout training, indicating that it may not be suitable for the architecture used. In contrast, tanh provided some improvement but still exhibited considerable deviation from the true ODE solutions. Figure \ref{fig:activation_functions} we can contrast the effect of different activation functions over the given \( t_{span} \) following the trends observed, both ReLU and tanh display irregular curves, stagnation, and overfitting while sigmoid follows a consistent downward trend.

Next, we varied the architecture by adjusting the number of hidden units and layers. Larger architectures with more hidden units captured the underlying dynamics better, leading to more accurate approximations. However, overly complex architectures introduced overfitting, emphasizing the need for balance in model complexity.

Finally, we experimented with different optimizers, and Adam emerged as the most efficient. It provided faster convergence compared to traditional stochastic gradient descent (SGD), particularly in capturing the intricate behavior of the system within the defined \( t_{span} \) of (0.0, 10.0).

Based on the analysis from Figure \ref{fig:ode_losses}, it is evident that the current configuration including the neural network architecture with reduced complexity, and the adoption of the sigmoid activation function, has proven to be highly effective in achieving a significantly reduced loss of $ \approx $ \textbf{2.15} in the search range. This represents a substantial improvement compared to the initial stages of training. The sigmoid activation function likely facilitated more gradual transitions between neuron activations, helping the network to capture intricate patterns and achieve finer-grained predictions, thereby contributing to the observed lower loss.

Additionally, reducing the number of hidden layers and making the network more shallow can have several beneficial effects. The computational burden is reduced by simplifying the network architecture, allowing for faster training iterations and potentially reducing the risk of overfitting the training data. This adjustment may have helped the network to generalize better to unseen data and focus on learning the essential features of the problem domain, thereby contributing to a lower loss.



\begin{figure}[h]
    \centering
    \begin{subfigure}[b]{0.49\textwidth}
        \centering
        \includegraphics[width=\textwidth]{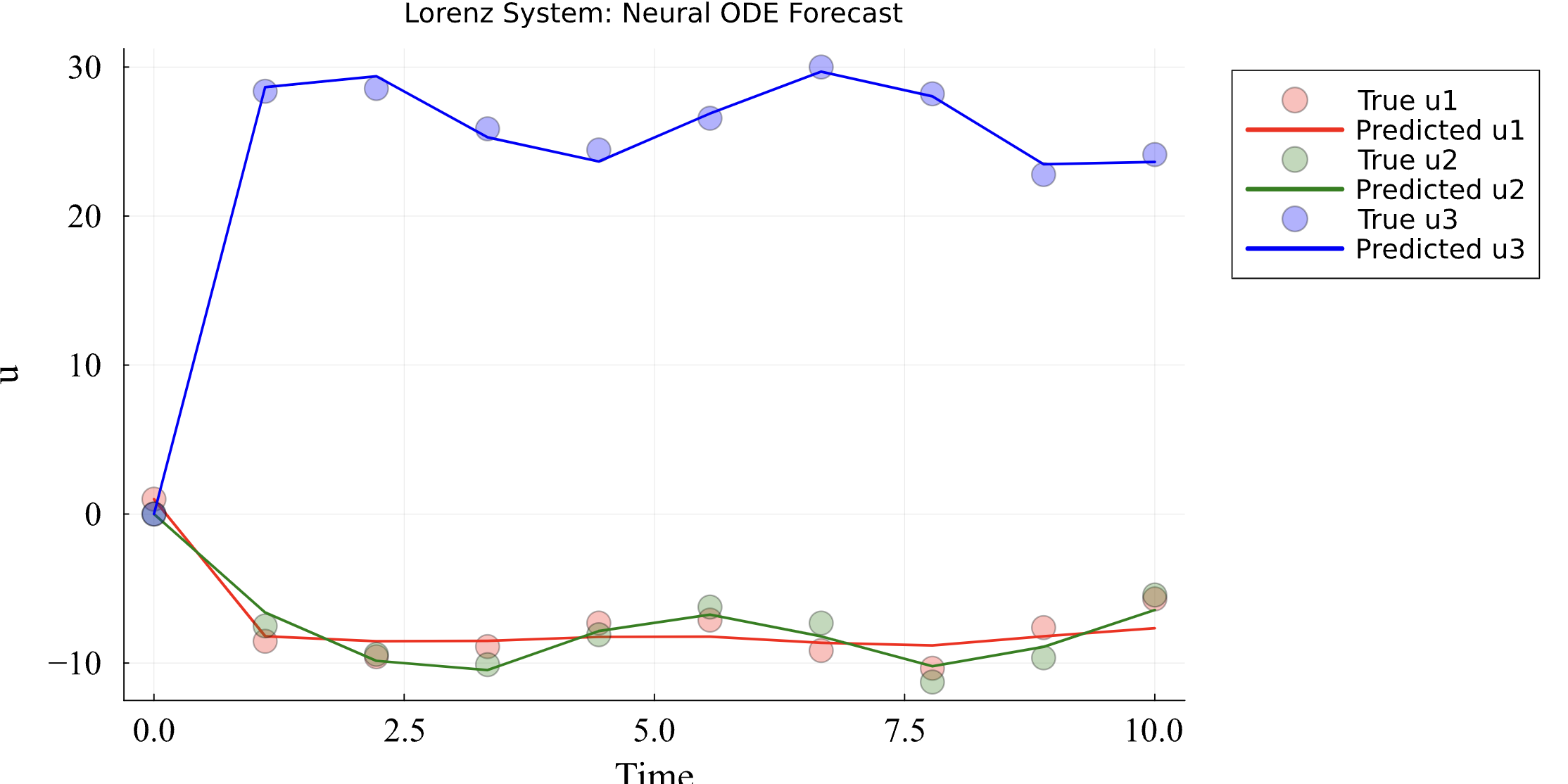}
        \caption{Training with a \( t_{span} \) 10}
        \label{fig:ode_training}
    \end{subfigure}
    \begin{subfigure}[b]{0.49\textwidth}
        \centering
        \includegraphics[width=\textwidth]{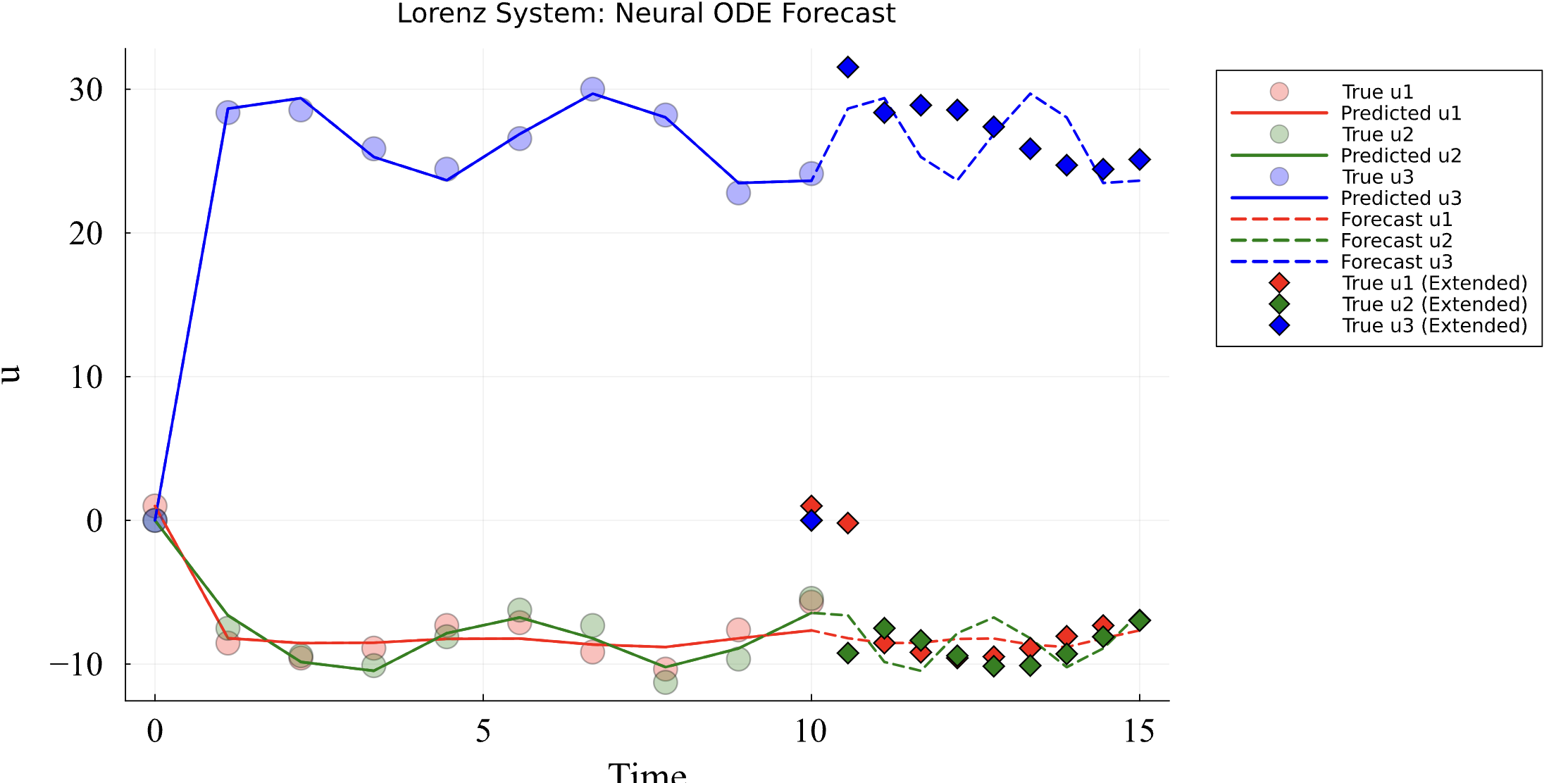}
        \caption{Extended forecast with \( t_{span} \) 15}
        \label{fig:neural_ode_forecast}
    \end{subfigure}
    \caption{Illustration of training and extended forecasting of the chaotic Lorenz system using Neural ODEs. The markers represent the true  \(u_1\), \(u_2\), and \(u_3\) and lines represent predicted values of the three variables over time.}
    \label{fig:both}
\end{figure}

Figure \ref{fig:ode_training} shows the training plot where the y-axis represents time, spanning from 0 to 10 units. The graph visually depicts the comparison between true data and predicted points, highlighting their consistent alignment at every time point. This alignment, achieved through meticulous training and optimization, is quantitatively reflected by a low loss value of 2.15. Such a minimal loss indicates that the neural ODE model effectively minimizes the discrepancy between predicted and actual data points across the entire time range.


Figure \ref{fig:neural_ode_forecast} depicts the forecast plot extending from time 0 to 15 units, the Neural ODE model's predictions are illustrated alongside the true data. Here, while the model demonstrates accurate alignment with the true data at certain points, it also diverges at others. This divergence is indicative of the model's ability to capture the underlying dynamics up to a certain extent, beyond which its predictive accuracy diminishes. Despite these discrepancies, the model's overall performance can still be evaluated based on its ability to generalize across the extended time interval, highlighting both its strengths and limitations in forecasting the system's behavior over an extended duration.

In summary, the combination of the sigmoid activation function, Adam optimizer with an appropriate learning rate of 0.01, and a streamlined neural network architecture has synergistically contributed to the significant reduction in loss observed in Table \ref{tab:hp-ode}, the same data can be observed as an exponential downward decrease in loss as shown in Figure \ref{fig:ode_losses}. \\

\subsection{Universal Differential Equations}


Table \ref{tab:hp-ude} displays the range of hyperparameters used in the UDE experiment. From previous findings of the Neural ODE experiment, we utilize the proven hyperparameters, activation function \textit{sigmoid}, hidden units (25), and epochs (50000), to obtain optimal training results.

\begin{table}[h]
	\caption{Range of hyper-parameters on training data (UDEs)}
	\centering
	\begin{tabular}{lll}
		\toprule
		\cmidrule(r){1-2}
		Hyperparameter   &Value   &Search Range    \\
		\midrule
          Activation Function    &sigmoid      &ReLU, sigmoid \\
          Learning Rate    &0.01      & 0.01, 0.001 \\
          Optimization Solver    &Adam, BFGS     &Adam, RAdam, BFGS \\
          Hidden units    &25      &25, 50 \\
          Number of Epochs    &50000      &10000 - 50000 \\
          Loss    &0.0257      &0.0257, 4.156, 10.5  \\
		\bottomrule
	\end{tabular}
	\label{tab:hp-ude}
\end{table}

Similar to neural ode, UDEs do not perform well when activation functions apart from sigmoid are employed, although they fare better results when compared to Neural ODEs, their overfitting and stagnation are still relatively high when compared to sigmoid activation function. 
Similarly, using a simple architecture of hidden layers less than 25 for example, produces better results with fewer resources when compared to a neural network with more than 50 or 100 hidden layers which only produces negligible improvements while consuming more time and compute resources.
 
\begin{figure}[h!]
    \centering
    \begin{subfigure}[b]{0.49\textwidth}
        \centering
        \includegraphics[width=\textwidth]{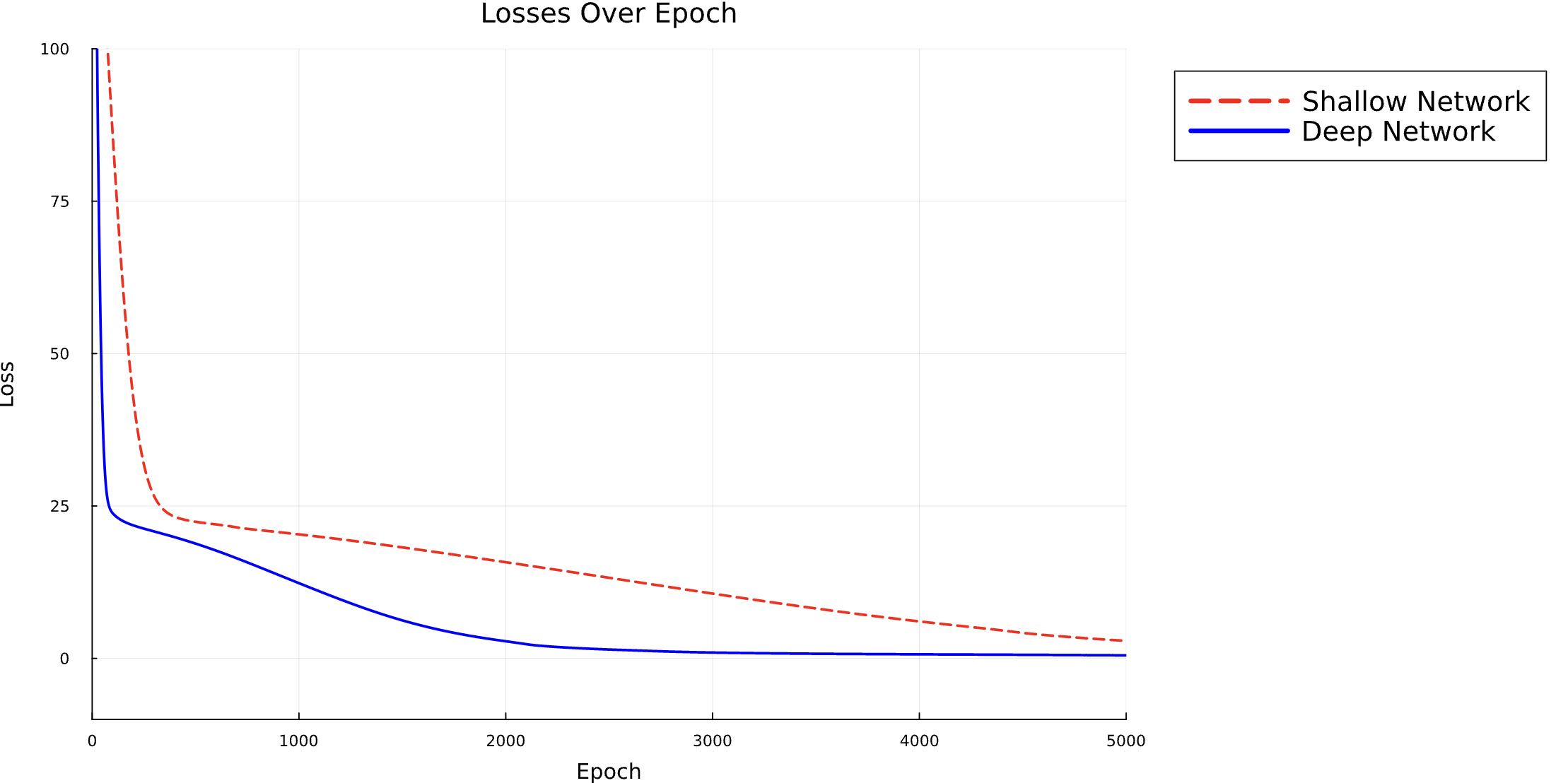}
        \caption{Deep vs Shallow network}
        \label{fig:deep_vs_shallow}
    \end{subfigure}
    \begin{subfigure}[b]{0.49\textwidth}
        \centering
        \includegraphics[width=\textwidth]{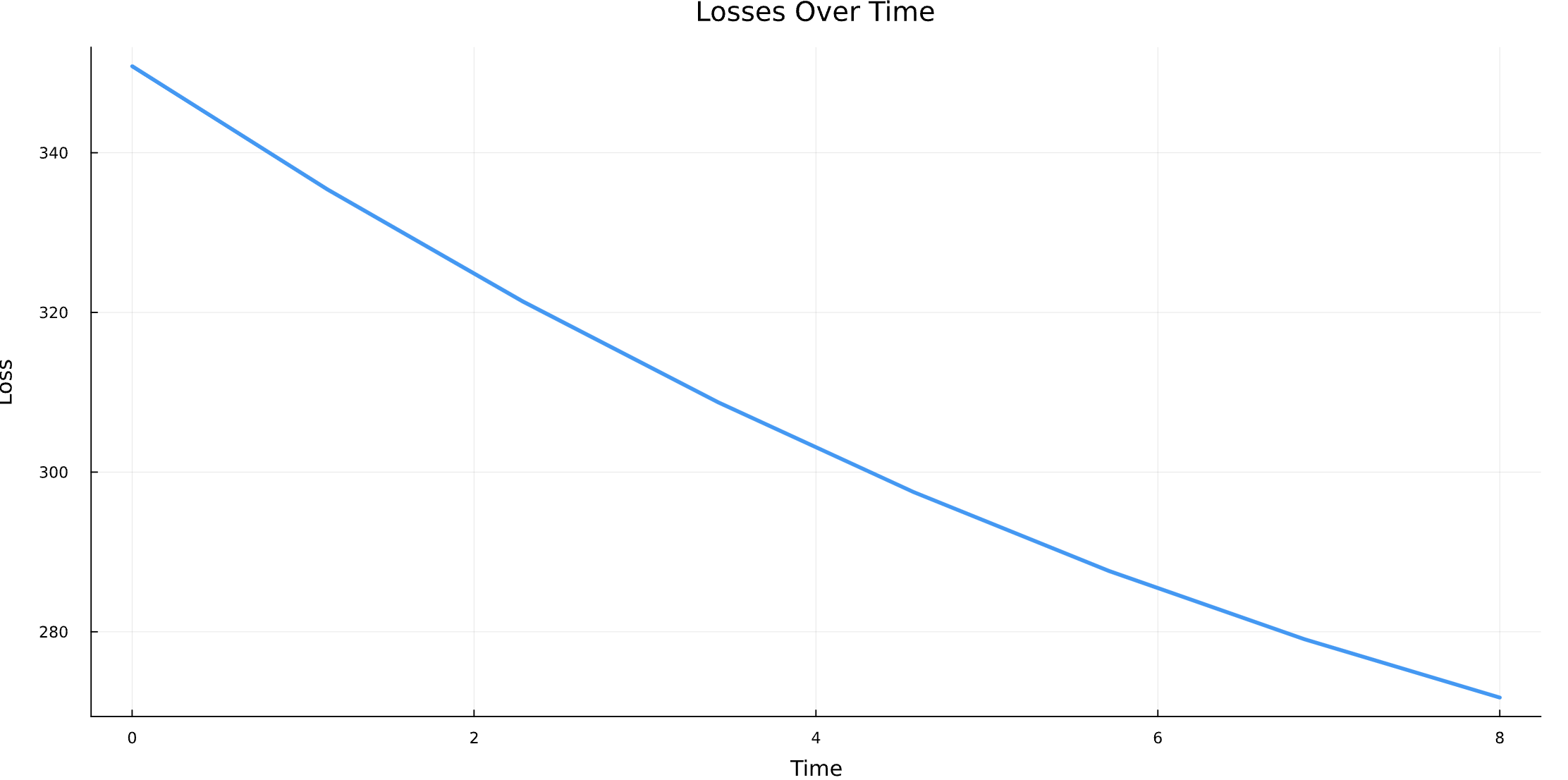}
        \caption{Loss over time}
        \label{fig:ude_loss_over_time}
    \end{subfigure}
    \caption{Comparison between architectures consisting of a deeper neural network vs a shallow neural network: There is some improvement in the loss when a network is trained on the deeper network, but this increases time significantly and does not provide substantial results when compared to a shallow network.}
    \label{fig:both}
\end{figure}

Figure \ref{fig:ude_loss_over_time} depicts the loss over time, from these findings it is clear that with the optimized parameter set, UDEs excel in identifying and recovering the missing terms, thereby reducing the loss to a minimal value. 
However, there was potential for further optimization. To achieve even better accuracy and further minimize the loss, we ran the UDE problem with a different optimizer. Specifically, we can refine the solution obtained from the initial prediction using the BFGS optimizer. This approach aims to enhance accuracy and reduce the loss even more effectively. 

By leveraging the BFGS optimizer, known for its efficiency in handling non-linear optimization problems, we can fine-tune the parameters to achieve a more precise model. The process involves re-evaluating the solution with the new optimizer, which iteratively adjusts the parameters to minimize the loss function further. This refinement step played a crucial in ensuring the UDE model reaches its optimal performance reducing the loss to $ \approx 0.0257$, thereby accurately representing the underlying system dynamics.

Comparing the line graphs of deep and shallow neural networks in Figure \ref{fig:deep_vs_shallow}, we observe the impact of increasing the number of hidden layers from 2 to 3 and expanding the dimensions to 100 in the third layer. The plot reveals a trend of decreasing points with a stagnation period between epochs 2000 onwards. Toward the end of 48000 epochs, the loss decreases by a factor of 10  ($ \approx 0.00107 $ ) compared to the shallow neural network results. While this performance improvement is noteworthy, it significantly increases the training time. Given that the initial loss value is already quite small, the advantages of employing a deeper neural network are debatable. Furthermore, the diminishing returns in performance raise questions about the necessity and efficiency of deeper architectures for certain tasks, suggesting that the benefits might not always justify the increased complexity and computational resources required. \\ 

\begin{figure}[h!]
    \centering
    \begin{subfigure}[b]{0.49\textwidth}
        \centering
        \includegraphics[width=\textwidth]{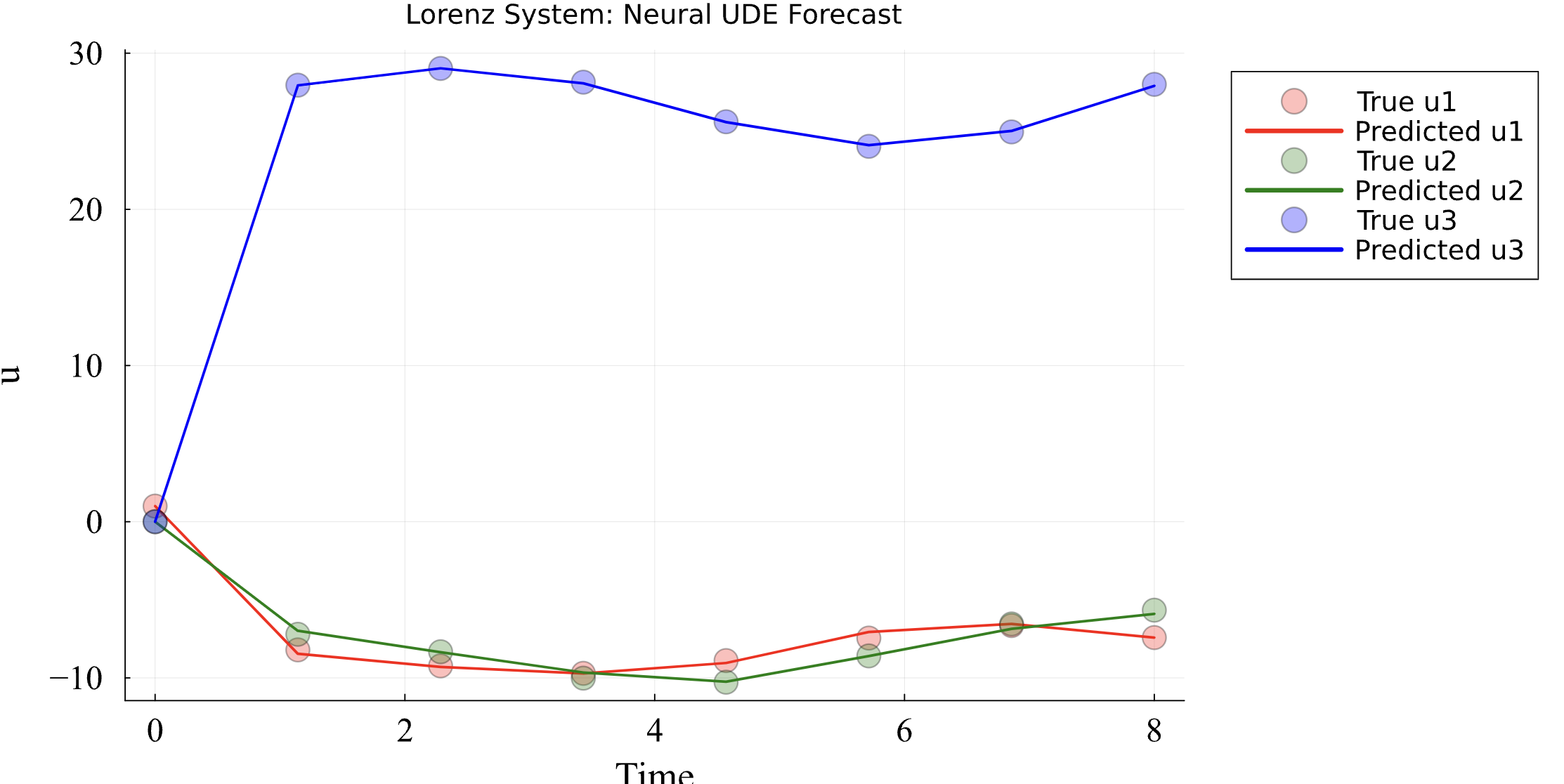}
        \caption{Training by recovery of terms}
        \label{fig:ude_training}
    \end{subfigure}
    \begin{subfigure}[b]{0.49\textwidth}
        \centering
        \includegraphics[width=\textwidth]{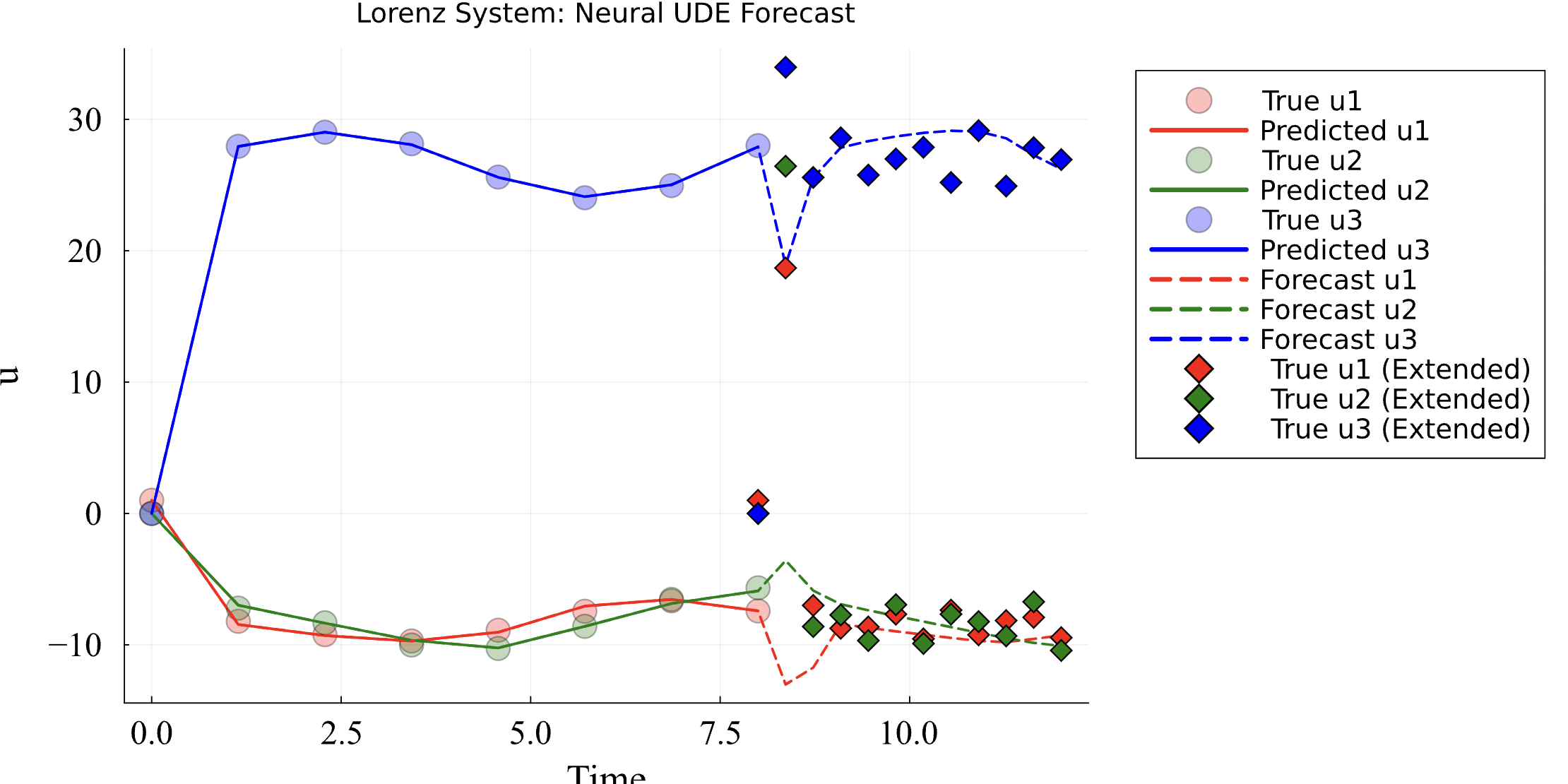}
        \caption{Forecast with \( t_{span} \) 15}
        \label{fig:ude_forecast}
    \end{subfigure}
    \caption{a) demonstrates the ability of the model to recover missing terms from the equations and white b) performs accurate predictions over increased time.}
    \label{fig:both}
\end{figure}

Figure \ref{fig:ude_training} depicts the plot of a Universal Differential Equation (UDE) training plot, the y-axis represents time, extending from 0 to 8 units. The graph demonstrates an exceptional alignment between true data and predicted points, where they meet precisely at every time point. This fit is a result of optimized parameter findings, which significantly reduced the loss to an exceptionally low value of 0.0257. The optimized parameters ensure that the model not only aligns perfectly with the true data within the observed interval but also exemplifies the potential of UDEs to achieve superior predictive performance. \\

Figure \ref{fig:ude_forecast} depicts the forecast plot for the trained UDE model, for an extended period from 0 to 15 units, the predictions closely align with the true data for the majority of the interval, showcasing the model's strong predictive capability. However, beyond the timespan, the prediction begins to diverge from the true data, highlighting a limitation of the model. This deviation suggests that while the UDE model excels within the observed time range, its ability to generalize and maintain accuracy diminishes over extended periods. This underscores the importance of continuous parameter optimization and model refinement to enhance long-term predictive performance in complex systems.

The successful application of Universal Differential Equations (UDEs) in this study has demonstrated their significant potential for accurately modelling complex systems with partially known dynamics. Achieving a remarkably low loss of $ \approx $ \textbf{0.0257} as depicted in Figure \ref{fig:ude_loss_over_time}, this underscores the robustness and efficacy of UDEs in integrating neural networks with traditional differential equations to capture intricate system behaviours and recover missing terms. This result highlights the enhanced accuracy and flexibility of UDEs, making them a powerful tool for scientific machine learning.

Furthermore, the efficiency of UDEs in learning from data and refining model parameters suggests their broad applicability across various scientific and engineering domains. Achieving such a low loss through advanced optimizers like BFGS indicates that UDEs can be further optimized to improve predictive capabilities and provide more reliable simulations. As research in this area progresses, UDEs are poised to advance scientific understanding and solve complex real-world problems, driving innovation and discovery in fields such as climate modelling, biological systems, and engineering design. \\

\begin{figure}[h!]
    \centering
        \centering
        \includegraphics[width=0.60\textwidth]{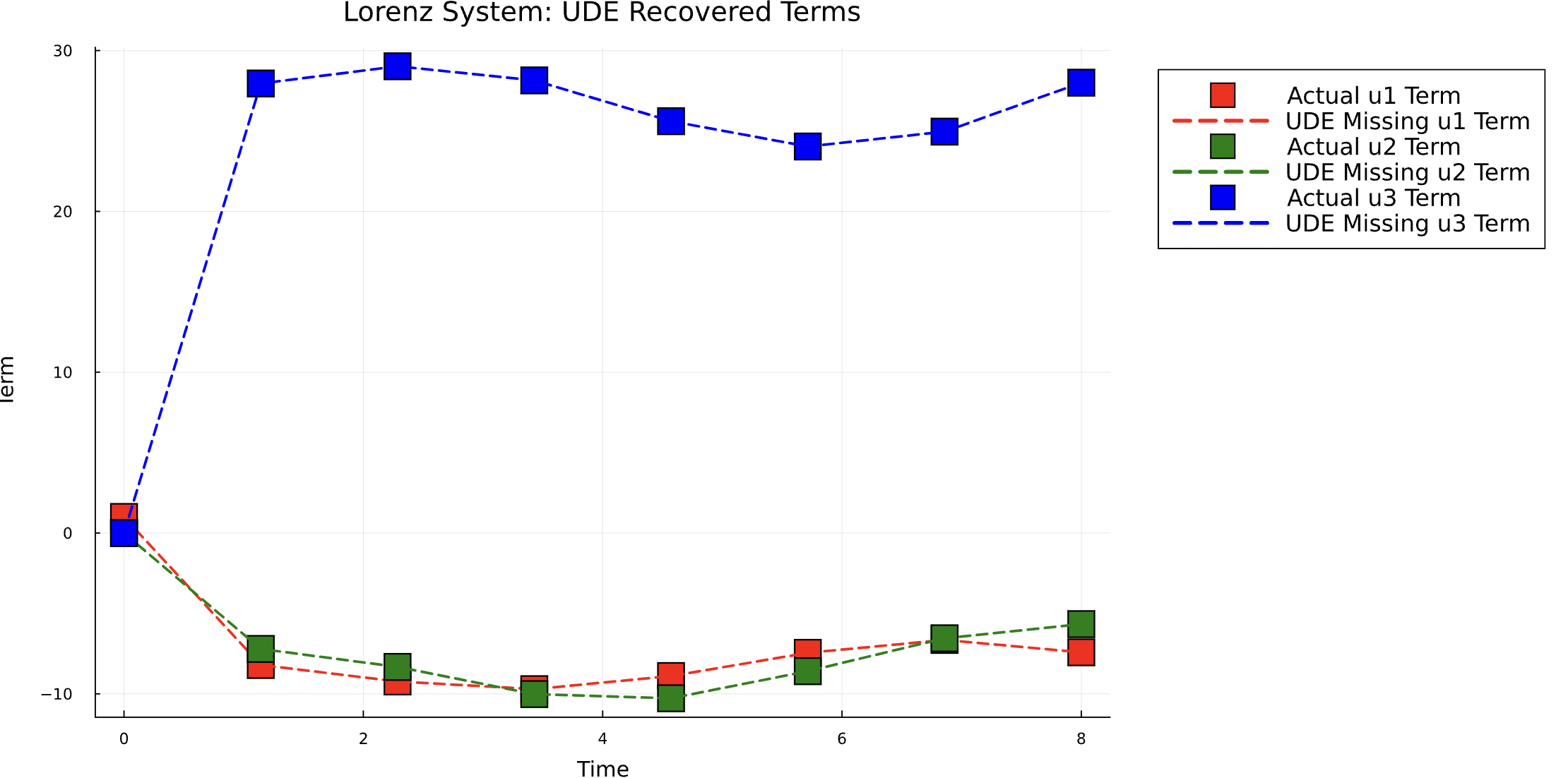}
        \caption{Recovery of missing terms: The UDE model is able to recover all of the missing terms when compared to the original data, this is showcased by each dotted line perfectly coinciding with the square blocks.}
        \label{fig:missing_terms}
\end{figure}

Figure \ref{fig:missing_terms} shows the plot of the missing terms compared to the actual terms in the Lorenz system and provides valuable insights into the behavior of the UDE model. The actual terms, plotted with circles, represent the true dynamics of the Lorenz system. The UDE missing terms, represented by dashed lines, indicate the corrections or adjustments made by the UDE model to match the true dynamics.

The near-linear nature of the UDE missing terms with slight downward trajectories across u1, u2, and u3 suggests that the neural network is primarily compensating for linear discrepancies in the original Lorenz system. These corrections are essential for aligning the UDE model's predictions with the true dynamics of the system. The consistency of the adjustments implies that the neural network has identified underlying linear trends that were not initially captured and is applying these corrections uniformly over time. This behaviour is crucial for improving the accuracy and robustness of the UDE model, particularly in chaotic systems like the Lorenz attractor where small errors can lead to significant deviations over time. \\

\subsection{Effect of noise}

\begin{figure}[h!]
    \centering
    \begin{subfigure}[b]{0.49\textwidth}
        \centering
        \includegraphics[width=\textwidth]{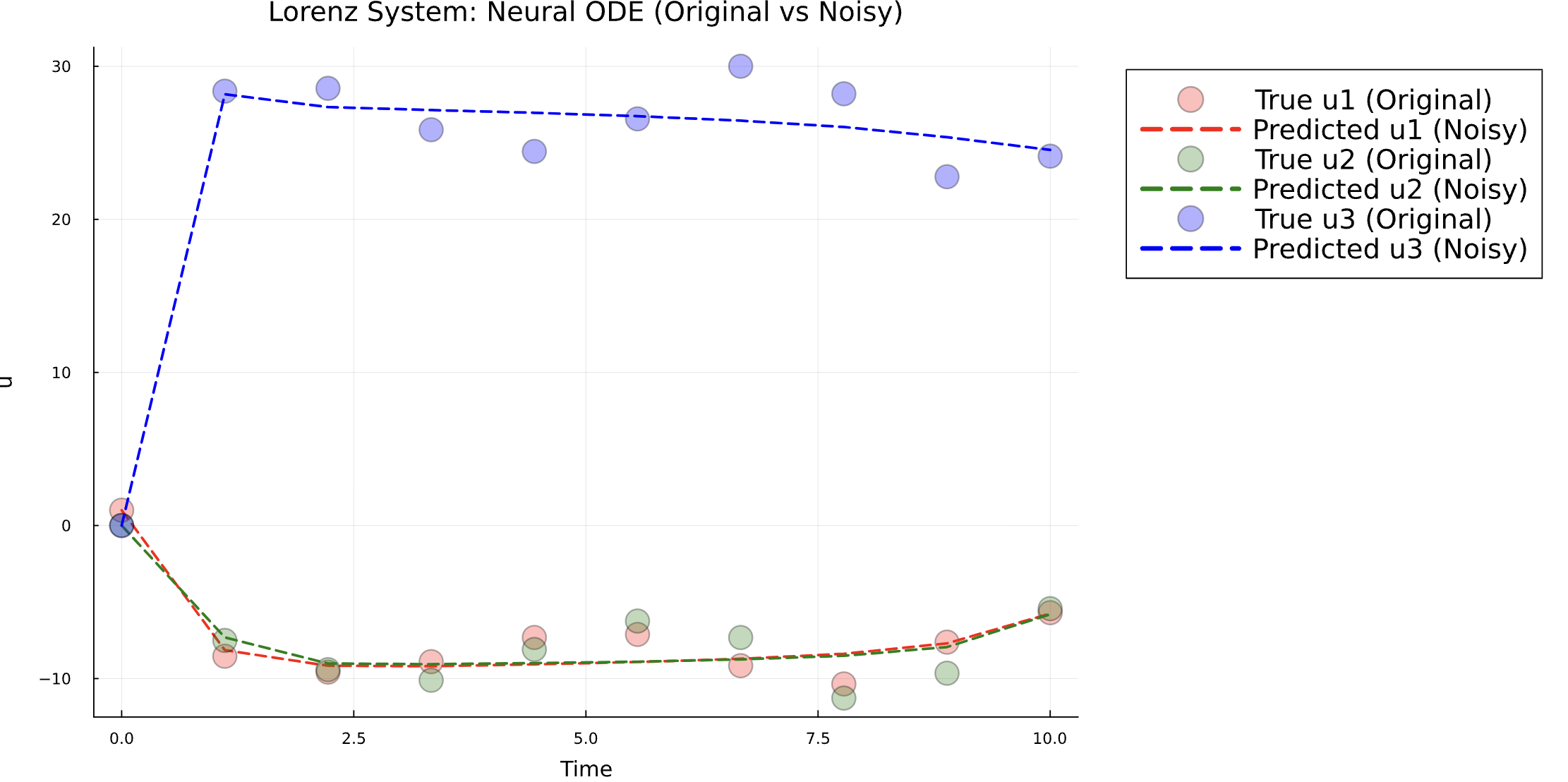}
        \caption{Neural ODE}
        \label{fig:ode_noisy}
    \end{subfigure}
    \begin{subfigure}[b]{0.49\textwidth}
        \centering
        \includegraphics[width=\textwidth]{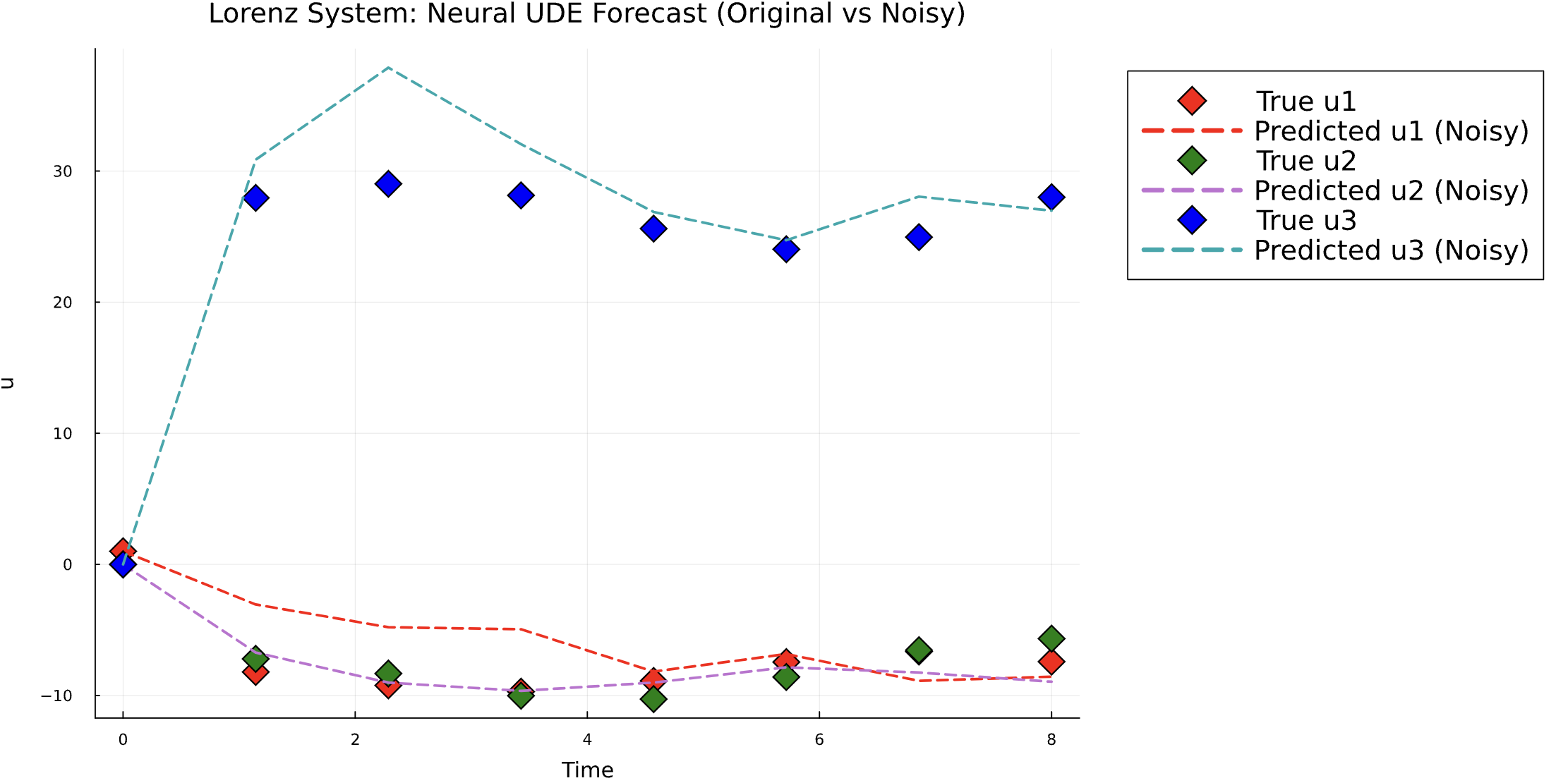}
        \caption{Neural UDE}
        \label{fig:ude_noisy}
    \end{subfigure}
    \caption{Although both models fail to produce accurate predictions when noise is introduced, UDE seems to capture more dynamics whereas neural ODE follows a mostly linear line.}
    \label{fig:both}
\end{figure}

The phenomenon observed in figures \ref{fig:ode_noisy} and \ref{fig:ude_noisy}, where the prediction markers for \(u_1\), \(u_2\), and \(u_3\) do not perfectly coincide with the ground truth data, can be attributed to the sensitivity of neural ODEs and UDEs to noise in the training data. During training, the neural network learns to approximate the underlying dynamics of the system based on the provided data. However, when noise is introduced into the training data (in this case, by adding Gaussian noise with a noise level of 0.1 to the original ODE data), the network must learn not only the true dynamics but also accommodate the noise. This can lead to overfitting, where the model captures the noise characteristics rather than the true underlying dynamics, resulting in poor generalization to unseen data or even to the same data when noise is present. The breakdown of predictions indicates that the model is not robust to the noise introduced during training, which is a common challenge in machine learning models dealing with differential equations.



\subsection{Forecasting breakdown points for the SciML models}

\begin{figure}[h!]
    \centering
    \begin{subfigure}[b]{0.49\textwidth}
        \centering
        \includegraphics[width=\textwidth]{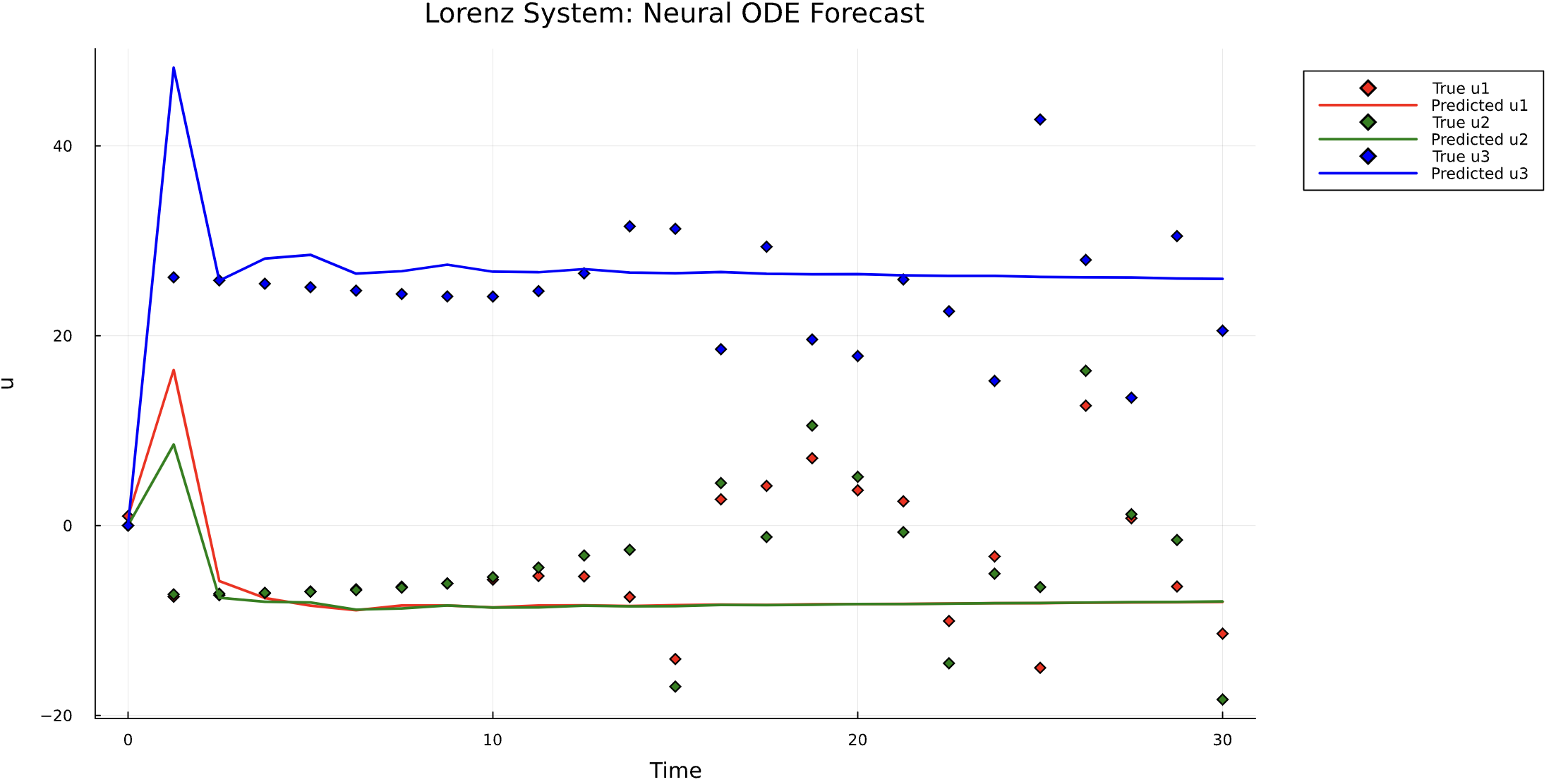}
        \caption{Neural ODE}
        \label{fig:ode_break}
    \end{subfigure}
    \begin{subfigure}[b]{0.49\textwidth}
        \centering
        \includegraphics[width=\textwidth]{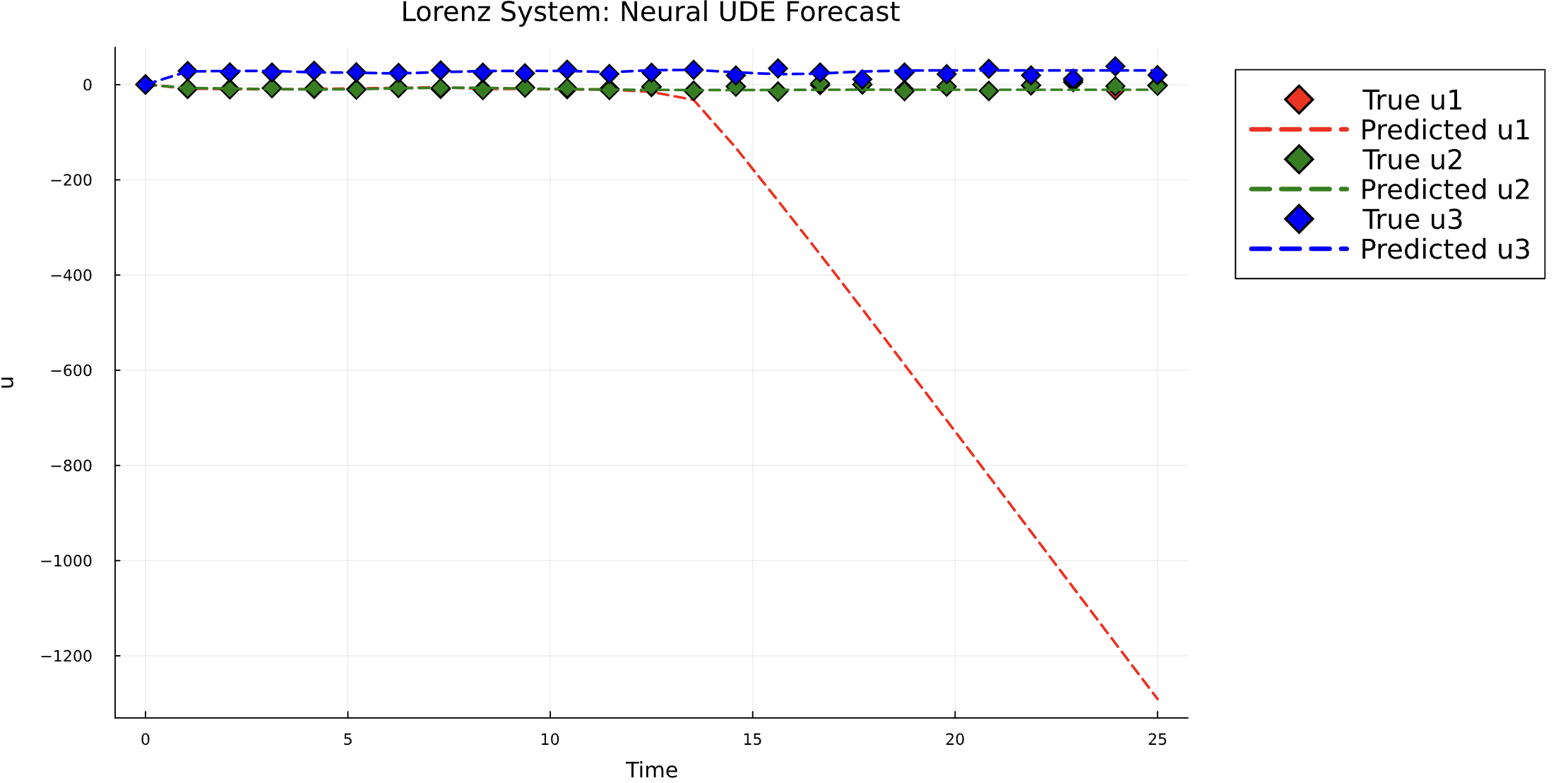}
        \caption{Neural UDE}
        \label{fig:ude_break}
    \end{subfigure}
    \caption{Limitations in modeling the chaotic system, neural ode fails to capture the underlying physics beyond the initial data points. A similar observation is recorded for UDE where prediction \(u_1\) completely breaks}
    \label{fig:both}
\end{figure}

 Despite the accurate training and prediction of both models, they come with their limitations. As shown by Figure \ref{fig:ode_break} the Neural ODE model beyond a certain point stops capturing the underlying physics and starts stagnating into a straight line indicating a need for continuous model validation and potential enhancements for long-term predictions. On the other hand in the case of UDE as shown by Figure \ref{fig:ude_break}, prediction for \(u_1\) completely breaks down towards a high negative value suggesting a lack of reliability for extended prediction. Future work will focus on improving the forecasting performance of these models by integrating symbolic regression to discover the symbolic formulations of recovered terms.


\section{Discussion and Conclusion}
We were able to approximate the underlying data very well through a trained Neural ODE. The combination of the sigmoid activation function, Adam optimizer with an appropriate learning rate, and
a streamlined neural network architecture has synergistically contributed to the significant reduction in loss observed. The Neural ODE model forecasts well from  0 to 15 units but breaks down after that. This divergence is indicative of the model’s
ability to capture the underlying dynamics up to a certain extent, beyond which its predictive accuracy diminishes.

UDEs excel in identifying and recovering the missing terms, thereby reducing the loss to a minimal value of ~ 0.0257. The efficiency of UDEs in learning from data and refining model parameters suggests their broad applicability across various scientific and engineering domains. Similar to the Neural ODE, the UDE model forecasts well till a timespan of 15 units. Beyond the timespan of 15 units, the prediction begins to diverge from the true data, highlighting a
limitation of the model. This deviation suggests that while the UDE model excels within the observed time range, its
ability to generalize and maintain accuracy diminishes over extended periods. 

\begin{table}[h]
	\caption{A preview of different parameters and results in the comparison of neural ODE and UDE models.}
        \vspace{0.1em}
	\centering
	\begin{tabular}{lll}
		\toprule
		\cmidrule(r){1-2}
		Parameter   &Neural ODE   &UDE    \\
		\midrule
          Prediction ability &\checkmark &\checkmark \\
          Forecast ability beyond training data
          &60\%  &66.7\%  \\
          Optimization solver    &Adam      &Adam + BFGS \\
          Final loss    &2.15      &0.0257 \\
		\bottomrule
	\end{tabular}
	\label{tab:preview}
\end{table}

Table \ref{tab:preview} summarizes the performance benchmarks for both experiments. Notably, both Neural ODEs and UDEs are capable of predicting outcomes up to 66\% beyond their training data. While Neural ODEs can be trained to achieve single-digit loss values, UDEs demonstrate superior learning capabilities, reducing the loss to below 0.05. This highlights their enhanced ability to generalize and produce more accurate predictions. While both Neural ODEs and UDEs can capture and predict complex dynamics over shorter intervals effectively, their accuracy and reliability can deteriorate over extended time spans. This underscores the need for continuous model validation and potential adjustments or enhancements to improve long-term forecasting
capabilities. 

 As Scientific Machine Learning methods are investigated, a lot of focus needs to be paid to forecasting. Most of the studies in the literature are aimed towards predictions. While the predictive power of SciML methods has been reliably demonstrated, as we show in the study, there are still doubts over their reliable forecasting abilities. In future work, we will modify the SciML models to ensure a better forecasting performance. We also plan to apply symbolic regression to the recovered terms to discover the symbolic formulations of the terms recovered by Neural ODEs and UDEs.

\bibliographystyle{unsrtnat}
\bibliography{references}  






\end{document}